\documentclass[preprint,12pt]{elsarticle}
\usepackage[named]{algo}
\usepackage{graphics}
\usepackage{graphicx}
\usepackage{stfloats}
\usepackage{amssymb}
\usepackage{amsmath}
\usepackage{amsfonts}
\usepackage{amsthm}
\usepackage{subfigure}
\usepackage{float}
\usepackage{pifont}
\usepackage{setspace}
\usepackage{url}

\newcommand{\Section}{\section}
\newcommand{\SubSection}{\subsection}
\newtheorem{theorem}{Theorem}
\newtheorem{lemma}{Lemma}

\newcommand{\eg}{{\it e.g.}}
\newcommand{\ie}{{\it i.e.}}
\DeclareMathOperator{\sign}{sign}
\DeclareMathOperator{\prox}{prox}

\def\urltilda{\kern -.15em\lower .7ex\hbox{\~{}}\kern .04em}
\def\urldot{\kern -.10em.\kern -.10em}
\def\urlhttp{http\kern -.10em\lower -.1ex\hbox{:}\kern -.12em\lower 0ex\hbox{/}\kern -.18em\lower 0ex\hbox{/}}

\begin{document}
\begin{frontmatter}

\title{Solving OSCAR regularization problems by proximal splitting algorithms}

\author[IST_IT]{Xiangrong Zeng \corref{cor1}}
\ead{Xiangrong.Zeng@lx.it.pt}

\author[IST_IT]{M\'{a}rio A. T. Figueiredo \corref{cor2}}
\ead{mario.figueiredo@lx.it.pt}

\cortext[cor1]{Corresponding author}
\cortext[cor2]{Principal corresponding author}

\address[IST_IT]{Instituto de Telecomunica\c{c}\~{o}es, Instituto Superior T\'{e}cnico, 1049-001, Lisboa, Portugal.}


\begin{abstract}
The OSCAR ({\it octagonal selection and clustering algorithm for regression})  regularizer consists of a $\ell_1$ norm plus
a pair-wise $\ell_{\infty}$ norm (responsible for its grouping behavior) and was  proposed to encourage group sparsity in scenarios where the groups are a priori unknown. The OSCAR regularizer has a non-trivial proximity operator, which limits its applicability. We reformulate this regularizer as a
weighted sorted $\ell_1$ norm, and propose its {\it grouping proximity operator} (GPO) and {\it approximate proximity operator} (APO),
thus making state-of-the-art proximal splitting algorithms (PSAs) available to solve inverse problems with OSCAR regularization.
The GPO is in fact the APO followed by additional grouping and averaging operations,
which are costly in time and storage, explaining the reason why algorithms with APO are much faster than that with GPO.
The convergences of  PSAs with GPO are guaranteed since GPO is an exact proximity operator. Although convergence of PSAs with APO is may not be guaranteed, we have experimentally found that APO behaves similarly to GPO when the regularization parameter of the pair-wise $\ell_{\infty}$ norm is set to an appropriately small value. Experiments on recovery of group-sparse signals (with unknown groups)
show that PSAs with APO are very fast and accurate.
\end{abstract}


\end{frontmatter}

\Section{Introduction}\label{sec:intro}

In the past few decades, linear inverse problems have attracted a lot of attention in a wide range of areas,
 such as statistics, machine learning, signal processing, and compressive sensing, to name a few.
The typical forward model is
\begin{equation}\label{linearmodel}
 {\bf y}={\bf A}{\bf x} +\bf {n},
\end{equation}
where  ${\bf y}\in\mathbb{R}^{m}$ is the measurement vector, ${\bf x}\in\mathbb{R}^{n}$ the original signal
to be recovered, ${\bf A}\in\mathbb{R}^{m\times n}$ is a known sensing matrix,
and ${\bf n}\in\mathbb{R}^{m}$ is noise (usually assumed to be white and Gaussian).
In most cases of interest, ${\bf A}$ is not invertible (\eg, because $m<n$),
making (\ref{linearmodel}) an ill-posed problem (even in the absence of noise),
which can only be addressed by using some form of regularization or
prior knowledge about the unknown ${\bf x}$.
Classical regularization formulations seek solutions of problems of the form
\begin{equation}\label{regularationmodel1}
\min_{\bf x} f({\bf x}) + \lambda\, r({\bf x})
\end{equation}
or one of the equivalent (under mild conditions) forms
\begin{equation}\label{regularationmodel2}
\min_{\bf x} r({\bf x}) \:\mbox{ s.t.} \: f({\bf x}) \leq \varepsilon \;\;  \mbox{ or } \;\;
\min_{\bf x} f({\bf x}) \:\mbox{ s.t.} \: r({\bf x}) \leq \epsilon,
\end{equation}
where, typically, $f({\bf x})=\tfrac{1}{2} \left\| {\bf y}-{\bf A}{\bf x} \right\|_2^2$ is the data-fidelity term (under a white Gaussian noise assumption), $r ({\bf x})$ is the regularizer that
enforces certain properties on the target solution, and $\lambda$, $\varepsilon$, and $\epsilon$ are
non-negative parameters.

A type of prior knowledge that has been the focus of much recent attention (namely with the advent of compressive sensing -- CS -- \cite{candes2006compressive}, \cite{donoho2006}) is sparsity, \ie, that a large fraction of the components of ${\bf x}$ are zero \cite{EladBook}.  The ideal regularizer encouraging
solutions with the smallest possible number of non-zero entries is $r_{\ell_0} ({\bf x}) = \left\|{\bf x}\right\|_0$ (which corresponds to the number of non-zero elements of ${\bf x}$), but the resulting problem is of combinatorial nature and known to be NP-hard \cite{candes2005decoding},  \cite{natarajan1995sparse}. The LASSO ({\it least absolute shrinkage and selection operator}) \cite{tibshirani1996regression} uses the $\ell_1$ norm, $r_{\mbox{\tiny LASSO} }
\left({\bf x}\right) = \left\|{\bf x}\right\|_1 = \sum_i |x_i|$, 
is arguably  the most popular sparsity-encouraging regularizer.  The  $\ell_1$  norm  can be seen as the tightest convex approximation of the $\ell_0$ ``norm" (it's not a norm) and, under conditions that are object of study in CS
\cite{candes2006stable}, yields the same solution. Many variants of these regularizers have been proposed, such as $\ell_p$ (for $p\in[0,1]$) ``norms"  \cite{chartrand2008,chen2010convergence}, and reweighted $\ell_1$ \cite{candes2008enhancing,wipf2010iterative,zhang2012reweighted}
and $\ell_2$ norms \cite{daubechies2004iterative,wipf2010iterative,chartrand2008iteratively}.

\SubSection{Group-sparsity-inducing Regularizers}

In recent years, much attention has been paid not only to the sparsity of solutions but also to the structure of this sparsity, which may be relevant in some problems and which provides another avenue for inserting prior knowledge into the problem.  In particular, considerable interest has been attracted by
group sparsity \cite{yuan2005model}, block sparsity \cite{eldar2009block}, or more general structured sparsity \cite{Bach2012}, \cite{huang2011learning}, \cite{micchelli2010regularizers}.
A classic model for group sparsity is the {\it group LASSO} (gLASSO) \cite{yuan2005model},
where the regularizer is the so-called $\ell_{1,2}$ norm \cite{qin2012structured},
\cite{liu2010fast} or the $\ell_{1,\infty}$ norm \cite{qin2012structured},
\cite{mairal2010network}, defined as
$r_{\mbox{\tiny gLASSO}}\left({\bf x}\right) = \sum_{i=1}^s \left\|{\bf x}_{g_i}\right\|_2$ and $\sum_{i=1}^s\left\|{\bf x}_{g_i}\right\|_{\infty}$,
respectively\footnote{Recall that $\|{\bf x}\|_{\infty} = \max\{|x_1|,|x_2|,...,|x_n|\}$}, where ${\bf x}_{g_i}$ represents the subvector of ${\bf x}$ indexed by
$g_i$, and  $g_i\subseteq \{1,...,n\}$ denotes the index set of the $i$-th group.
Different ways to define the
groups lead to overlapping or non-overlapping gLASSO.
Notice that if each group above is a singleton, then $r_{\mbox{\tiny gLASSO}}$
reduces to $r_{\mbox{\tiny LASSO}}$, whereas if $s=1$ and $g_1 = \{1,...,n\}$, then
$r_{\mbox{\tiny gLASSO}}({\bf x}) = \|{\bf x}\|_2$. Recently, the {\it sparse gLASSO} (sgLASSO) regularizer was proposed as
$r_{\mbox{\tiny sgLASSO}}\left({\bf x}\right) = \lambda_1 r_{\mbox{\tiny LASSO}}\left({\bf x}\right) +
\lambda_2 r_{\mbox{\tiny gLASSO}}\left({\bf x}\right) $,  where $\lambda_1$ and  $\lambda_2$ are non-negative parameters \cite{simon2012sparse}.
In comparison with gLASSO, sgLASSO not only selects groups,
but also individual variables within each group. Note that one of the costs of the possible advantages of
gLASSO and sgLASSO  over standard LASSO is the need to define {\it a priori} the structure of the groups.

In some problems, the components of ${\bf x}$ are known to be similar in value
to its neighbors (assuming that there is some natural neighborhood relation defined among the components of ${\bf x}$). To encourage this type of solution (usually in conjunction with
sparsity), several proposals have appeared, such as the elastic net \cite{zou2005regularization}, the fused LASSO (fLASSO) \cite{tibshirani2004sparsity}, {\it grouping pursuit} (GS) \cite{shen2010grouping},
and the {\it octagonal shrinkage and clustering algorithm for regression} (OSCAR) \cite{bondell2007simultaneous}. The elastic net regularizer is
$r_{\mbox{\tiny elast-net}}\left({\bf x}\right) = \lambda_1 \left\|{\bf x}\right\|_1 + \lambda_2 \left\|{\bf x}\right\|_2^2 $,
encouraging both sparsity and grouping \cite{zou2005regularization}.
The fLASSO regularizer is given by $r_{\mbox{\tiny fLASSO}}\left({\bf x}\right) = \lambda_1 \left\|{\bf x}\right\|_1 + \lambda_2 \sum_i |x_i - x_{i+1}| $, where
the {\it total variation} (TV) term (sum of the absolute values of differences) encourages consecutive variables to be similar; fLASSO is thus able to promote
both sparsity and smoothness. The GS regularizer 
is defined as  $r_{\mbox{\tiny GS}}\left({\bf x}\right) = \sum_{i<j}G\left(x_i - x_{j}\right)$, where $G\left( {\bf z}\right) = | {\bf z}|$, if
$| {\bf z}|\leq \lambda$, and $G\left( {\bf z}\right) = \lambda$, if $| {\bf z}|> \lambda$ \cite{shen2010grouping}; however, $r_{\mbox{\tiny GS}}$ is neither sparsity-promoting nor convex.
Finally, $r_{\mbox{\tiny OSCAR}}$ \cite{bondell2007simultaneous} has the form
$r_{\mbox{\tiny OSCAR}}\left({\bf x}\right) = \lambda_1 \left\|{\bf x}\right\|_1 + \lambda_2 \sum_{i<j}\max\left\{|x_i|,|x_j| \right\}$; due to $\ell_1$ term
and the pair-wise $\ell_{\infty}$ penalty, the components are encouraged to be
sparse and pair-wise similar in magnitude.

Other recently proposed group-sparsity  regularizers include the
adaptive LASSO (aLASSO) \cite{zou2006adaptive},
 where the regularizer is 
 $r_{\mbox{\tiny aLASSO}}\left({\bf x}\right) = \lambda \sum_i |x_i |/|\tilde{x}_i |^\gamma $,
 where ${\bf \tilde{x}}$
is an initial consistent estimate of ${\bf x}$, and $\lambda$ and $\gamma$ are positive parameters.
The {\it pairwise} fLASSO (pfLASSO \cite{petry2011pairwise}) is a variant of fLASSO, given by $r_{\mbox{\tiny pfLASSO}}\left({\bf x}\right) =  \lambda_1 \left\|{\bf x}\right\|_1 + \lambda_2 \sum_{i<j} | x_i - x_j |$,
is related to OSCAR, and extends fLASSO to cases where the variables have no natural ordering. Another variant is the {\it weighted} fLASSO (wfLASSO \cite{daye2009shrinkage}), given by
 $r_{\mbox{\tiny wfLASSO}}\left({\bf x}\right) =  \lambda_1 \left\|{\bf x}\right\|_1 + \lambda_2 \sum_{i<j} w_{ij}
\left( x_i - \mbox{sign} \left( \rho_{ij} \right) x_j \right)^2 $,
where $\rho_{ij}$ is the sample correlation between the $i$-th and $j$-th predictors,
and  $w_{ij}$ is a non-negative weight. Finally, the recent {\it graph-guided} fLASSO (ggfLASSO \cite{kim2009multivariate}) regularizer
is based on a graph $G = ({\bf V}, {\bf E})$, where ${\bf V}$ is the set of variable nodes and
${\bf E}\subseteq {\bf V}^2$ the set of edges:
$r_{\mbox{\tiny ggfLASSO}}\left({\bf x}\right) =  \lambda_1 \left\|{\bf x}\right\|_1 + \lambda_2 \sum_{(i,j) \in
 {\bf E}, i<j} w_{ij} | x_i - \mbox{sign} \left( r_{ij} \right) x_j |$,
where $r_{ij}$  represents the weight of the edge $\left(i, j\right) \in {\bf E}$;
if $r_{ij} = 1$, $r_{\mbox{\tiny ggfLASSO}}$ reduces to $r_{\mbox{\tiny fLASSO}}$, and the former can group variables with different signs through the 
assignment of $r_{ij}$, while the latter cannot. Some other graph-guided
group-sparsity-inducing regularizers have been proposed in \cite{yang2012feature}, 
and all this kind of regularizers needs a strong requirement -- the prior information on an undirected graph.  

\begin{figure}
\centering
		\includegraphics[width=0.8\columnwidth]{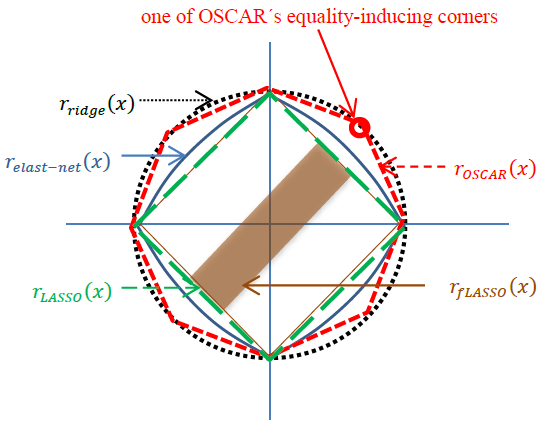}
		\caption{Illustration of $r_{\mbox{\tiny fLASSO}}$, $r_{\mbox{\tiny elast-net}}$, $r_{\mbox{\tiny LASSO}}$,
		$r_{\mbox{\tiny OSCAR}}$ and $r_{\mbox{\tiny ridge}}$
		which is the regularizer of ridge regression.}
		\label{fig:manyregs}
\end{figure}

For the sake of comparison, several of the above mentioned regularizers are illustrated in Figure \ref{fig:manyregs},
where the corresponding level curves (balls) are depicted; we also
plot the level curve of the classical ridge regularizer $r_{\mbox{\tiny ridge}} = \|{\bf x}\|_2^2$.
We can see that $r_{\mbox{\tiny OSCAR}}\left({\bf x}\right)$,
$r_{\mbox{\tiny elast-net}}\left({\bf x}\right)$, and $r_{\mbox{\tiny fLASSO}}\left({\bf x}\right)$ promote
both sparsity and grouping, but their grouping behaviors are clearly  different:
1) OSCAR encourages equality (in magnitude)
of each pair of variables, as will be discussed in detail in Section \ref{sec:OSCARAPOGPO};
2)  elastic net is strictly convex, but doesn't promote strict equality like OSCAR;
3) the total variation term in fLASSO can be seen to encourage sparsity in the differences between
each pair of successive variables, thus its recipe of grouping is to guide variables into the shadowed
region shown in Figure \ref{fig:manyregs}), which corresponds to
 $\sum_i | x_i - x_j | \leq \varsigma $ (where $\varsigma$ is a function of  $\lambda_2$).

As seen above, fLASSO, the elastic net, and OSCAR have the potential to be
used as regularizers when it is known that the components
of the unknown signal exhibit structured sparsity, but a group structure
is not {\it a priori} known. However, as pointed out in \cite{zhong2012efficient},
OSCAR outperforms the other two models in feature grouping. Moreover, fLASSO is not suitable
for grouping according to magnitude, since it cannot group positive and negative variables together, even if their magnitudes are similar; fLASSO also relies on a particular ordering of the variables, for which there may not always be a 
natural choice. Consequently, we will focus on the OSCAR regularizer in this paper.

In \cite{bondell2007simultaneous}, a costly quadratic programming approach was adopted
to solve the optimization problem corresponding to OSCAR. More recently, \cite{petry2011oscar}
solved OSCAR in a generalized linear model context; the algorithm therein proposed solves
a complicated constrained maximum likelihood problem in each iteration, which is
also costly. An efficient algorithm was proposed in
\cite{zhong2012efficient}, by reformulating  OSCAR as a quadratic problem and then
applying FISTA ({\it fast iterative shrinkage-thresholding algorithm}) \cite{beck2009fast}.
To the best of our knowledge, this is the currently fastest algorithm for OSCAR. In this paper,
we propose reformulating $r_{\tiny \mbox{OSCAR}}$ as a weighted and sorted $\ell_1$ norm,
and present an exact grouping proximity operator (termed GPO) of $r_{\tiny \mbox{OSCAR}}$
that is based on an projection step proposed in \cite{zhong2012efficient}
and an element-wise approximate proximal step (termed APO). We show that GPO
consists of APO and an additional grouping and averaging operation.
Furthermore, we use alternative state-of-the-art {\it proximal
splitting algorithms} (PSAs, briefly reviewed next)
to solve the problems involved by the OSCAR regularization.

\SubSection{Proximal Splitting and Augmented Lagrangian Algorithms}

In the past decade, several special purpose algorithms have been proposed to solve
optimization problems of the form \eqref{regularationmodel1} or \eqref{regularationmodel2}, in the context of linear inverse problems
with sparsity-inducing regularization. For example,  homotopy methods \cite{osborne2000lasso},
LARS \cite{efron2004least}, and StOMP \cite{donoho2006sparse} deal with these
problems through adapted active-set strategies; $\ell_1-\ell_s$ \cite{tsaig2007sparse}
and GPSR \cite{figueiredo2007gradient} are based on bound-constrained optimization.
Arguably, the standard PSA is the so-called {\it iterative shrinkage/thresholding}
(IST) algorithm, or forward-backward splitting \cite{combettes2005signal}, \cite{hale2007fixed}, \cite{figueiredo2003algorithm}, \cite{daubechies2004iterative}, \cite{figueiredo2005bound}. The fact that IST tends to be slow, in particular if matrix ${\bf A}$ is poorly conditioned, has stimulated much research aiming at obtaining faster variants. In {\it two-step IST} (TwIST  \cite{bioucas2007new}) and in the {\it fast IST algorithm} (FISTA \cite{beck2009fast}), each iterate depends on the two previous ones,
rather than only on the previous one (as in IST). TwIST and FISTA have been shown
(theoretically and experimentally) to be considerably faster than standard IST. Another  strategy  to  obtain  faster  variants of  IST consists in using more aggressive step sizes; this is the case of SpaRSA
({\it sparse reconstruction by separable approximation}) \cite{wright2009sparse}, which was also shown to clearly outperform standard IST.

The {\it split augmented Lagrangian shrinkage algorithm} (SALSA \cite{afonso2010fast,afonso2011augmented})
addresses unconstrained optimization problems based on variable splitting \cite{courant1943variational}, \cite{wang2008new}. The idea is to transform the unconstrained problem into a constrained one via variable splitting, and then tackle this constrained problem using the {\it alternating direction method of multipliers} (ADMM) \cite{boyd2011distributed,Eckstein1992}. Recently, a preconditioned version of ADMM (PADMM), which is a primal-dual scheme, was proposed in  \cite{chambolle2011first}. ADMM has a close relationship \cite{setzer2009split} with Bregman and split-Bregman methods (SBM) \cite{yin2008bregman,goldstein2009split,osher2011fast,langer2010analysis,goldstein2009split},
which have been recently applied to imaging inverse problems \cite{zhang2012reweighted,cai2009split,setzer2010deblurring}.
The six state-of-the-art methods mentioned above -- FISTA, TwIST, SpaRSA, SBM,
ADMM, and PADMM will be considered in this paper, in the context of numerical experiments.

\SubSection{Contribution of the Paper}
The contributions of this paper are two-fold: 1) We reformulate the OSCAR regularizer as
a weighted sorted $\ell_1$ norm,  and propose the APO and GPO; this is the main
contribution, which makes solving the corresponding optimization problems more convenient;
2) we study the performance of the six state-of-the-art algorithms FISTA, TwIST, SpaRSA, SBM,
ADMM and PADMM, with APO and GPO, in solving problems with OSCAR regularization.

\SubSection{Organization of the Paper}

The rest of the paper is organized as follows. Section II describes OSCAR, and its GPO and APO.
Due to limitation of space, Section III only details three of six algorithms: FISTA,
SpaRSA, and PADMM. Section IV reports experimental results and Section V concludes the paper.

\SubSection{Terminology and Notation}

We denote vectors and matrices by lower and upper case bold letters,
respectively. The $\ell_p$ norm of a vector ${\bf x}\in\mathbb{R}^{n}$ is $\|{\bf x}\|_p=\left(\sum_{i=1}^n |x_i|^p\right)^{1/p}$.
We denote as $|{\bf x}|$ the vector with the absolute values of the elements of ${\bf x}$ and as ${\bf x}\odot {\bf z}$ the element-wise multiplication of two vectors of the same dimension. Finally, $\sign(v) = 1$, if $v\geq 0$ and
$\sign(v) = -1$, if $v<0$; if the argument is a vector, the $\sign$ function is understood in a component-wise fashion.

We new briefly review some elements of convex analysis used below.
Let $\mathcal{H}$ be a real Hilbert space with inner product $\langle\cdot,\cdot\rangle$ and norm $\left\|\cdot\right\|$. Let $f:\mathcal{H}\rightarrow \mathbb{R} \cup \{+\infty\}$ be a function and
$\Gamma$ be the class of all lower semi-continuous, convex, proper functions
(not equal to $+\infty$ everywhere). The (Moreau) proximity operator
\cite{Moreau1962,combettes2005signal} of $f\in \Gamma$ is defined as
\begin{equation}\label{proximityoperator}
\prox_{f} \left( {\bf v}\right) =
\arg\min_{{\bf x}\in \mathcal{H}} \left( f\left( {\bf x} \right)
+ \frac{1}{2} \left\|{\bf x}-{\bf v}\right\|^2\right).
\end{equation}
If $f ( {\bf x} )= \iota_C ( {\bf x})$, the indicator function of a nonempty closed
convex set $C$ (\ie, $\iota_C({\bf x}) = 0$, if ${\bf x}\in C$, and $\iota_C({\bf x}) = +\infty$, if ${\bf x}\not\in C$), then  $\prox_{\lambda f} ( {\bf x})$ is the projection of ${\bf x}$ onto $C$. If $f$ is the $\ell_1$ norm, then (\ref{proximityoperator}) is the well-known soft thresholding:
\begin{equation} \label{softthresholding}
\mbox{soft}( {\bf v} , \lambda ) = \mbox{sign} ({\bf v}) \odot \max \{ |{\bf v}| - \lambda, 0 \}.
\end{equation}

The conjugate of $f \in \Gamma$ is defined by
$f^*( {\bf u}) = \sup_{\bf x} \left\langle {\bf x}, {\bf u}  \right\rangle - f \left( {\bf x}\right)$,
such that $f^*\in \Gamma$, and $f^{**}=f$. The so-called Moreau's identity states that \cite{combettes2005signal},
\begin{equation} \label{moerauidentity}
{\bf x} = \prox_{\lambda f } ( {\bf x} ) + \lambda \, \prox_{f^*/\lambda } \left( {\bf x}/\lambda \right).
\end{equation}

\Section{OSCAR and Its Proximity Operator}\label{sec:OSCARAPOGPO}

\SubSection{OSCAR} \label{subsec:OSCAR}

\begin{figure} [htbp]
	\centering
		\includegraphics[width=0.75\columnwidth]{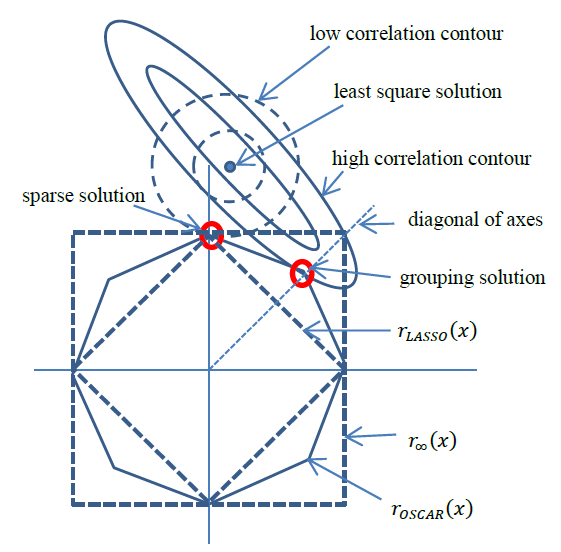}
	\caption{Illustration of the $\ell_2$ term and  $r_{\mbox{\tiny OSCAR}} \left( {\bf x}\right)$
	in the $n=2$ case, where $r_{\infty}\left({\bf x}\right)$ is the $\ell_\infty$ norm.
	In this example, for the same least square solution $({\bf A}^T{\bf A})^{-1}{\bf A}^T{\bf y}$,
	the high correlation contour is more likely to hit the equality-inducing (grouping) vertex, whereas the
    low correlation contour prefers the sparsity-inducing vertex. }
	\label{fig:OSCAR}
\end{figure}

The OSCAR criterion is given by
\begin{equation}\label{OSCAR}
\min_{\bf x} \frac{1}{2} \left\| {\bf y} - {\bf A}{\bf x} \right\|_2^2 +
\underbrace{\lambda_1 \left\|{\bf x} \right\|_1 + \lambda_2 \sum_{i<j}
\max \left\{ |x_i |,  |x_j |\right\}}_{r_{\mbox{\tiny OSCAR}} \left( {\bf x}\right)},
\end{equation}
where ${\bf A}\in\mathbb{R}^{m\times n}$,  ${\bf x}\in\mathbb{R}^{ n}$ \cite{bondell2007simultaneous}.
In (\ref{OSCAR}), the $\ell_2$ term seeks data-fidelity, while the regularizer
$r_{\mbox{\tiny OSCAR}} \left( {\bf x}\right)$ consists of an $\ell_1$ term
(promoting sparsity) and a pairwise $\ell_\infty$ term ($(n(n-1))/2$ pairs in total)
encouraging  equality (in magnitude) of each pair of elements $\bigl( |x_i| , |x_j|\bigr)$. Thus, $r_{\mbox{\tiny OSCAR}} \left( {\bf x}\right)$ promotes both sparsity and grouping.
Parameters $\lambda_1$ and $\lambda_2$ are nonnegative constants controlling
the relative weights of the two terms. If $\lambda_2=0$, \eqref{OSCAR} becomes the
LASSO, while if $\lambda_1=0$,  $r_{\mbox{\tiny OSCAR}}$ behaves only like a pairwise
$\ell_\infty$ regularizer. Note that, for any choice of $\lambda_1, \lambda_2\in\mathbb{R}^+$,
$r_{\mbox{\tiny OSCAR}} \left( {\bf x}\right)$ is convex and its ball is octagonal if $n=2$. The 8 vertices of this octagon can be divided into two categories: four sparsity-inducing
vertices (located on the axes) and the other four vertices which are equality-inducing.
Figure \ref{fig:OSCAR} depicts the a data-fidelity term and $r_{\mbox{\tiny OSCAR}} \left( {\bf x}\right)$,
illustrating its possible effects.

As discussed in Section \ref{sec:intro}, compared with gLASSO, the OSCAR doesn't require
a pre-specification of group structure; compared with fLASSO, it doesn't depend on a
certain ordering of the variables; compared with the elastic net, it has the equality-inducing
capability. All these features make OSCAR a convenient regularizer in many applications.

A fundamental building block for using OSCAR is its proximity operator,
\begin{equation}\label{POofOSCAR}
\prox_{r_{\mbox{\tiny OSCAR}}} \left( {\bf v}\right) =
\arg\min_{{\bf x}\in \mathbb{R}^n} \left( r_{\mbox{\tiny OSCAR}} \left( {\bf x}\right)  + \frac{1}{2} \left\|{\bf x}-{\bf v}\right\|^2\right).
\end{equation}
Directly computing \eqref{POofOSCAR} seems untractable, thus we follow the
reformulation of $r_{\mbox{\tiny OSCAR}} \left( {\bf x}\right)$ introduced in
\cite{bondell2007simultaneous}, \cite{zhong2012efficient}.
Given some vector ${\bf x}$, let $\tilde{\bf x}$ result from sorting the components of ${\bf x}$ in decreasing order of magnitude (\ie $|\tilde{x}_1| \geq |\tilde{x}_2| \geq ... \geq |\tilde{x}_n|$) and  ${\bf P}({\bf x})$ be the corresponding
permutation matrix (with ties broken by some arbitrary but fixed rule),  that is,
$\tilde{\bf x} = {\bf P(x) \, x}$. Notice that ${\bf x} = \bigl({\bf P}({\bf x})\bigr)^T {\bf \tilde{x}}$, since the inverse of a permutation matrix is its transpose. With this notation, it is easy to see that
\begin{equation}\label{weightedsortedl1norm}
r_{\mbox{\tiny OSCAR}} \left( {\bf x}\right) = \|{\bf w} \odot {\bf \tilde{x}} \|_1 = \sum_{i=1}^{n} w_i |\tilde{x}_i|,
\end{equation}
where ${\bf w} $ is a weight vector with elements given by
\begin{equation}\label{weightvector}
w_i=\lambda_1+ \lambda_2 \left(n-i\right),\;\;\mbox{for}\;i=1,...,n.
\end{equation}
It is worth noting that the representation in \eqref{weightvector} is only unique
if the sequence of values of the elements of $\tilde{\bf x}$ is strictly decreasing.
Notice also that the choice (\ref{weightvector}) is always correct,
regardless of there existing or not repeated absolute values.

In order to use \eqref{weightedsortedl1norm}--\eqref{weightvector} to compute
(\ref{POofOSCAR}), we first introduce the following lemma.
\vspace{0.3cm}
\begin{lemma}
\label{lem:lemma1}
Consider ${\bf v} \in \mathbb{R}^{n }$, ${\bf P}({\bf v})$, and
${\bf \tilde{v}} = {\bf P(v)\, v}$ as defined above.
Let \begin{equation}\label{getmustar}
{\bf u}^* = \arg\min_{{\bf u}\in \mathbb{R}^n} \left( \|{\bf \boldsymbol{\pi}} \odot {\bf u} \|_1 +
\frac{1}{2} \left\|{\bf u}-{\bf \tilde{v}} \right\|^2 \right)
\end{equation}
where ${\bf \boldsymbol{\pi}}\in \mathbb{R}_+^n$ is a positive weight vector. If
\begin{equation}\label{keycondition}
|\tilde{v}_i| - {\pi}_i \geq  |\tilde{v}_{i+1}| -
{\pi}_{i+1}, \;\; \mbox{for}\; i=1,\dots,n-1,
\end{equation}
then ${\bf u}^*$  satisfies $|u^*_i|\geq |u^*_{i+1}|$, for  $i=1,2,...,n-1$.
Moreover, if \eqref{keycondition} is satisfied with ${\bf \boldsymbol{\pi}} = {\bf w}$, with ${\bf w}$ as given in \eqref{weightvector},  then $\bigl({\bf P}({\bf v})\bigr)^T {\bf u}^* = \prox_{r_{\mbox{\tiny OSCAR}}} ( {\bf v})$.
\end{lemma}
\vspace{0.3cm}

\begin{proof}
\label{the:proof1}
The minimization in (\ref{getmustar}) is in fact a set of $n$
decoupled minimizations, each of the form
\[\begin{split}
u_i^* & = \arg\min_{u} \pi_i |u| + \frac{1}{2}( u - \tilde{v}_i )^2\\
& = \mbox{soft}(\tilde{v}_i,\pi_i)\\
      &= \mbox{sign}(\tilde{v}_i) \max\{|\tilde{v}_i| - \pi_i,0\}.
\end{split}
\label{mustarinproof}
\]
If $|\tilde{v}_{i} | - {\pi}_i \geq  |\tilde{v}_{i+1} | -
{\pi}_{i+1},$ then
\[
\max\{|\tilde{v}_i| - \pi_i,0\} \geq \max\{|\tilde{v}_{i+1}| - \pi_{i+1},0\},
\]
that is, $|u_i^*| \geq |u_{i+1}^*| $. Now, if $|u_i^*| \geq |u_{i+1}^*|$, then
${\bf P}({\bf u}^*) = {\bf I}$, thus $\tilde{\bf u}^* = {\bf u}^*$, thus $
\|{\bf \boldsymbol{\pi}} \odot \tilde{\bf u}^* \|_1 = \|{\bf \boldsymbol{\pi}} \odot {\bf u}^* \|_1$. Thus, if  \eqref{keycondition} is satisfied with ${\bf \boldsymbol{\pi}} = {\bf w}$, then the term $\|{\bf \boldsymbol{\pi}} \odot {\bf u}^* \|_1  = r_{\mbox{\tiny OSCAR}} \left({\bf u}^* \right)$ (according to \eqref{weightedsortedl1norm} and recalling that ${\bf u}^* = \prox_{r_{\mbox{\tiny OSCAR}}} \left( \tilde{\bf v}\right)$),  thus $ \bigl({\bf P}({\bf v})\bigr)^T {\bf u}^* = \prox_{r_{\mbox{\tiny OSCAR}}} \left( {\bf v}\right)$.
\end{proof}
\vspace{0.3cm}

Notice that condition \eqref{keycondition}, with
 ${\bf \boldsymbol{\pi}} = {\bf w}$, is equivalent to
\begin{equation} \label{piisw}
|\tilde{v}_i| - |\tilde{v}_{i+1}| \geq w_i - {w_{i+1}}, \;\; \mbox{for}\; i=1,\dots,n-1.
\end{equation}
Unfortunately, \eqref{piisw} is a strong requirement on ${\bf \tilde{v}}$ and, therefore,
generally, $\bigl({\bf P}({\bf v})\bigr)^T {\bf u}^* \neq \prox_{r_{\mbox{\tiny OSCAR}}}({\bf v})$.
 Next, we introduce a {\it grouping proximity operator} (GPO), which is the exact proximity operator
for any ${\bf {v}}$.

\SubSection{Grouping Proximity Operator (GPO) of $r_{\mbox{\tiny OSCAR}}\left({\bf x}\right)$ }  \label{subsec:GPO}
We next propose the exact GPO of $r_{\mbox{\tiny OSCAR}}$, based on the
projection step proposed in \cite{zhong2012efficient}. For the sake of completeness,
we briefly introduce the projection step here and then present
the GPO, leveraging Lemma \ref{lem:lemma1}.
Begin by noting that, since $r_{\mbox{\tiny OSCAR}}({\bf x}) = r_{\mbox{\tiny OSCAR}}(|{\bf x}|) $
and, for any ${\bf x},{\bf v} \in \mathbb{R}^n$ and ${\bf u} \in \{-1,1\}^n$,
\[
\|\mbox{sign}({\bf v})\odot |{\bf x}| - {\bf v}\|_2^2  \leq \| {\bf u} \odot |{\bf x}| - {\bf v}\|_2^2,
\]
we have that
\begin{equation} \label{signfunction}
\mbox{sign}\left(\prox_{r_{\mbox{\tiny OSCAR}}} \left( {\bf v}\right) \right) = \mbox{sign}({\bf v}).
\end{equation}
Consequently, we have
\begin{equation}\label{xtarfromx}
\prox_{r_{\mbox{\tiny OSCAR}}}({\bf v}) = \mbox{sign}\left({\bf v}\right) \odot \prox_{r_{\mbox{\tiny OSCAR}}}(|{\bf v}|),
\end{equation}
showing that there is no loss of generality in assuming ${\bf v} \geq 0$, \ie, the fundamental step in computing $\prox_{r_{\mbox{\tiny OSCAR}}}({\bf v})$ is
obtaining
\begin{equation}\label{a_with_v}
\begin{split}
{\bf a} &= \arg\min_{{\bf x}\in \mathbb{R}_+^n} \left( r_{\mbox{\tiny OSCAR}}\left({\bf x}\right) +
\frac{1}{2} \left\|{\bf x}- |{\bf v}|\right\|^2 \right).
 \end{split}
\end{equation}
Furthermore, both terms in the objective function in
\eqref{a_with_v} are invariant under permutations of the components of
the vectors involved; thus,
denoting
\begin{equation}\label{b_with_vsort}
{\bf b} = \arg\min_{{\bf x}\in \mathbb{R}_+^n } \left( r_{\mbox{\tiny OSCAR}}\left({\bf x}\right)
+ \frac{1}{2} \left\|{\bf x}- |{\bf \tilde{v}}|\right\|^2 \right),
\end{equation}
where (as defined above) $ \tilde{\bf v} = {\bf P}({\bf v})\; {\bf v}$,  allows writing
\begin{equation} \label{a_and_b}
{\bf a} = \bigl({\bf P}({\bf v})\bigr)^T{\bf b},
\end{equation}
showing that there is no loss of generality in assuming that
the elements of ${\bf v}$ are sorted in non-increasing magnitude.
As shown in the following theorem (from \cite{zhong2012efficient}),
${\bf b}$  has several important properties.

\begin{theorem}
\label{the:theorem2}
(\cite{zhong2012efficient}, Theorem 1 and Propositions 2, 3 and 4) Let ${\bf b}$ be  defined as in \eqref{b_with_vsort}; then:

\begin{itemize}
	\item [(i)] For  $i=1,2,...,n-1$, $b_i\geq b_{i+1}$;
	moreover, $(|\tilde{v}_p| = |\tilde{v}_{q}|) \Rightarrow ( b_p= b_q)$.
	\item [(ii)] The property of ${\bf b}$ stated in (i) allows
   writing it as
   \[{\bf b} =
[{ b}_1 ... { b}_n]^T
= \left[{ b}_{s_1}...
{ b}_{t_1}...{ b}_{s_j}...
{ b}_{t_j}...
{ b}_{s_l}...{ b}_{t_l} \right]^T,
\]
where $b_{s_j} = \cdots = b_{t_j}$ is the $j$-th group of consecutive
equal elements of ${\bf b}$ and there are $ 1\leq l \leq n$ such
groups. For the $j$-th group, the common optimal value is \begin{equation}\label{average}
{ b}_{s_j}=\cdots={ b}_{t_j} =
\max \left\{{ \bar{v}}_j - \bar{w}_j, 0 \right\}
\end{equation}
where
\begin{equation} \label{zbarandwbar1}
 { \bar{v}}_j =  \frac{1}{\vartheta_j}\sum_{i = s_j}^{t_j}|\tilde{v}_{i}|,
\end{equation}
is the $j$-th group average (with ${\vartheta_j} = t_j - s_j + 1$ denoting its number of components) and
\begin{equation}  \bar{w}_j  =
 \frac{1}{\vartheta_j} \sum_{i = s_j}^{t_j} w_{i}
 = \lambda_1 + \lambda_2 \left(n - \frac{s_j + t_j}{2}\right).\label{zbarandwbar2b}
\end{equation}
	\item [(iii)] For the $j$-th group, if ${ \bar{v}}_j - \bar{w}_j \geq 0 $, then
	there is no integer $c \in \left\{s_j,\cdots,t_j-1\right\}$
	such that \[
    \sum_{i=s_j}^c |\tilde{v}_i|- \sum_{i=s_j}^c w_i > 	
    \sum_{i=c+1}^{t_j} |\tilde{v}_i|- \sum_{i=c+1}^{t_j} w_i.
\] Such a group is called {\rm coherent}, since it cannot be split
into two groups with different values that respect (i).
\end{itemize}
 \end{theorem}
\vspace{0.3cm}

The proof of Theorem \ref{the:theorem2} (except the second equality \eqref{zbarandwbar2b}, which can be
derived from \eqref{weightvector}) is given in \cite{zhong2012efficient}, where
an algorithm was also proposed to obtain the optimal ${\bf b}$.
That algorithm equivalently divides the indeces of $|{\bf \tilde{v}}|$ into
groups and performs averaging within each group (according to \eqref{zbarandwbar1}),
obtaining a vector that we denote as ${\bf \bar{v}}$; this operation of grouping and averaging is denoted in this paper as
\begin{equation}\label{groupandaverage}
\bigl(\bar{\bar{{\bf v}}},\bar{\bar{{\bf w}}}\bigr) = \mbox{GroupAndAverage}\left(|{\bf \tilde{v}}|,{\bf w}\right),
\end{equation}
where
\begin{equation}\label{newzsort}
\bar{\bar{{\bf v}}} = [
\underbrace{{ \bar{v}}_1,\ldots,{ \bar{v}}_1}_{\vartheta_1 \; \mbox{\footnotesize components}}
\ldots \underbrace{{ \bar{v}}_j\ldots{ \bar{v}}_j}_{\vartheta_j  \; \mbox{\footnotesize components}}
\ldots \underbrace{{ \bar{v}}_l\ldots{ \bar{v}}_l}_{\vartheta_l \; \mbox{\footnotesize components}} ]^T,
\end{equation}
and
\begin{equation}\label{newwsort}
\bar{\bar{{\bf w}}} = [
\underbrace{{ \bar{w}}_1,\ldots,{ \bar{w}}_1}_{\vartheta_1 \; \mbox{\footnotesize components}}
\ldots \underbrace{{ \bar{w}}_j\ldots{ \bar{w}}_j}_{\vartheta_j  \; \mbox{\footnotesize components}}
\ldots \underbrace{{ \bar{w}}_l\ldots{ \bar{w}}_l}_{\vartheta_l \; \mbox{\footnotesize components}} ]^T,
\end{equation}
with the $\bar{v}_j$ as given in \eqref{zbarandwbar1} and
the $\bar{w}_j$ as given in \eqref{zbarandwbar2b}.
Finally, ${\bf b}$ is obtained as
\begin{equation}\label{equivalentform}
{\bf b} = \max(\bar{\bar{{\bf v}}}-\bar{\bar{\bf  w}}, 0 ).
\end{equation}

The following lemma, which is a simple corollary of Theorem \ref{the:theorem2},
indicates that condition \eqref{keycondition} is satisfied with $\boldsymbol{\pi} = \bar{\bar{\bf  w}}$.

\vspace{0.3cm}
\begin{lemma}
\label{lem:lemma2}
Vectors $\bar{\bar{{\bf v}}}$ and $\bar{\bar{\bf  w}}$ satisfy
\begin{equation}
\bar{\bar{v}}_i - \bar{\bar{w}}_i \geq  \bar{\bar{v}}_{i+1} -
\bar{\bar{w}}_{i+1}, \;\; \mbox{for}\; i=1,\dots,n-1.
\end{equation}
 \end{lemma}
\vspace{0.3cm}

Now, we are ready to give the following theorem for the GPO:

\vspace{0.3cm}
\begin{theorem}
\label{the:theorem3}
Consider ${\bf v} \in \mathbb{R}^{n }$ and a permutation matrix ${\bf P}({\bf v})$
such that the elements of ${\bf \tilde{v}} = {\bf P}({\bf v})\, {\bf v}$ satisfy
$|\tilde{v}_i| \geq |\tilde{v}_{i+1}|$, for $i=1,2,...,n-1$.
Let ${\bf a}$ be given by \eqref{a_and_b}, where ${\bf b}$ is given by
\eqref{equivalentform}; then ${\bf a}\odot \sign({\bf v}) =
\prox_{r_{\mbox{\tiny OSCAR}}} ( {\bf v})$.
\end{theorem}
\vspace{0.3cm}

\begin{proof}
\label{the:proof3}
According to Theorem \ref{the:theorem2}, ${\bf b}$ as given by \eqref{equivalentform} is the optimal solution of  \eqref{b_with_vsort}, then
${\bf a}\odot \sign({\bf v})$ is equal $\prox_{r_{\mbox{\tiny OSCAR}}} ( {\bf v})$,
according to $\eqref{xtarfromx}$ and $\eqref{b_with_vsort}$.
\end{proof}
\vspace{0.3cm}

\begin{figure} [h]
	\centering
		\includegraphics[width=1.0\columnwidth]{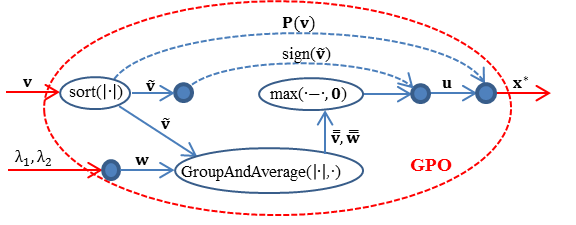}
	\caption{Procedure of computing GPO.  }
	\label{fig:GPO}
\end{figure}

The previous theorem shows that the GPO can be computed by the following algorithm
(termed $\mbox{OSCAR\_GPO}\left({\bf v}, \lambda_1, \lambda_2\right)$ where ${\bf v}$ is the input vector and $\lambda_1, \lambda_2$ are the parameters of
$r_{\mbox{\tiny OSCAR}}\left({\bf x}\right) $), and illustrated in Figure \ref{fig:GPO}.
\vspace{0.3cm}
\begin{algorithm}{$\mbox{OSCAR\_GPO}\left({\bf v},\lambda_1,\lambda_2\right)$}{
\label{alg:OSCARGPO}}
\textbf{input} ${\bf v}\in\mathbb{R}^{n}, \lambda_1, \lambda_2 \in \mathbb{R}_+$ \\
\textbf{compute}\\
     \ \ \ \ ${\bf w} = \lambda_1 + \lambda_2 \left( n - \left[1:1:n\right]^T \right)$\\
		 \ \ \ \ $\bigl({\bf \tilde{v}},{\bf P}({\bf v})\bigr) = \mbox{sort}\bigl( |{\bf v}| \bigr)$\\
		\ \ \ \ $\left(\bar{\bf \bar{\bf v}}, \bar{\bf \bar{\bf w}}\right) =
                           \mbox{GroupAndAverage}\left(|{\bf \tilde{v}}|,{\bf w}\right)$\\
		\ \ \ \ ${\bf u} = \mbox{sign}({\bf \tilde{v}})\odot \max(\bar{\bar{{\bf v}}}-\bar{\bar{\bf w}}, 0 )$\\
    \ \ \ \ ${\bf x}^* = \bigl({\bf P}({\bf v})\bigr)^T {\bf u} $\\
\qreturn ${\bf x}^*$
\end{algorithm}
\vspace{0.3cm}

GPO is the exact proximity operator of the OSCAR regularizer. Below, we
propose a faster approximate proximity operator.

\SubSection{Approximate Proximity Operator (APO) of $r_{\mbox{\tiny OSCAR}}\left({\bf x}\right)$ }\label{subsec:APO}
The GPO described in the previous subsection is the exact proximity operator of $r_{\mbox{\tiny OSCAR}}\left({\bf x}\right)$. In this subsection, we propose an
approximate version of GPO (named APO -- {\it approximate proximity operator}),
obtained by skipping the function $\mbox{GroupAndAverage}\left(\right)$. In
the same vein as the GPO, the APO of $r_{\mbox{\tiny OSCAR}}\left({\bf x}\right)$ is  illustrated as Figure 5, and the pseudo code of computing the APO, termed $\mbox{OSCAR\_APO}\left({\bf v},\lambda_1,\lambda_2\right)$,
is as follows.

\vspace{0.3cm}
\begin{algorithm}{$\mbox{OSCAR\_APO}\left({\bf v},\lambda_1,\lambda_2\right)$}{
\label{alg:OSCARAPO}}
\textbf{input} ${\bf v}\in\mathbb{R}^{n }, \lambda_1, \lambda_2 \in \mathbb{R}_+$ \\
\textbf{compute}\\
     \ \ \ \   ${\bf w} = \lambda_1 + \lambda_2 \left( n - \left[1:1:n\right]^T \right)$\\
		 \ \ \ \ $\bigl({\bf \tilde{v}},{\bf P}({\bf v})\bigr) = \mbox{sort}\bigl( |{\bf v}| \bigr)$\\
      \ \ \ \  ${\bf u} = \mbox{sign}({\bf \tilde{v}})\odot \max(\tilde{\bf v}-{\bf w}, 0 )$\\
    \ \ \ \ ${\bf x}^* = \bigl({\bf P}({\bf v})\bigr)^T {\bf u} $\\
\qreturn ${\bf x}^*$
\end{algorithm}
\vspace{0.3cm}
\begin{figure} [h]
	\centering
		\includegraphics[width=1.0\columnwidth]{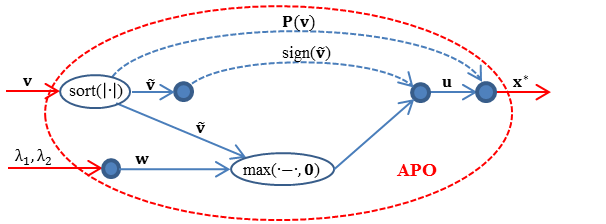}
	\caption{Procedure of computing APO.   }
	\label{fig:APO}
\end{figure}

APO can also be viewed as a group-sparsity-promoting variant of soft thresholding, obtained by removing (\ref{groupandaverage}) from the exact computation of the proximity operator of the OSCAR regularizer. With this alteration, \eqref{keycondition} is no longer guaranteed to be satisfied, this being the
reason why APO may not yield the same result as the GPO.
However, we next show that the APO is able to obtain results that are often as good as those produced by the GPO, although it is simpler and faster.

\SubSection{Comparisons between APO and GPO}

\begin{figure} [t]
	\centering
		\includegraphics[width=0.95\columnwidth]{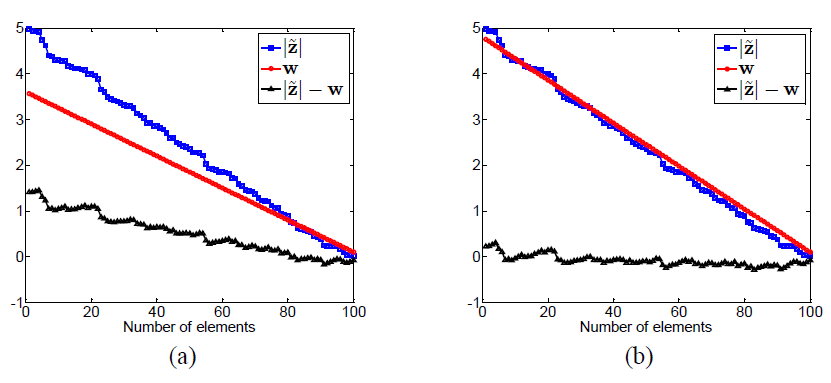}
	\caption{Relationships among $|\tilde{\bf z}|$, ${\bf w}$ and
	$|\tilde{\bf z}| - {\bf w}$ for different values of $\lambda_2$:
	(a) $\lambda_1=0.1$ and $\lambda_2=0.03$; (b) $\lambda_1=0.1$ and $\lambda_2=0.047$.}
	\label{fig:zsortandw}
\end{figure}

In this section, we illustrate the difference between APO and GPO on simple examples: let ${\bf z} \in \mathbb{R}^{100}$ contain random samples uniformly distributed in $\left[-5,5\right]$, and $\tilde{\bf z}={\bf P}({\bf z})\, {\bf z}$
be a magnitude-sorted version, \ie, such that $|\tilde{z}_i|>|\tilde{z}_{i+1}|$ for $i=1, \cdots, n-1$.
In Figure \ref{fig:zsortandw}, we plot $|\tilde{\bf z}|$, ${\bf w}$ (recall (\ref{weightedsortedl1norm})), and $|\tilde{\bf z}| - {\bf w}$, for different
values of $\lambda_2$; these plots show that, naturally, condition \eqref{keycondition} may not be satisfied and that its probability of
violation increases for larger values of $\lambda_2$.

\begin{figure}
	\centering
		\includegraphics[width=1.0\columnwidth]{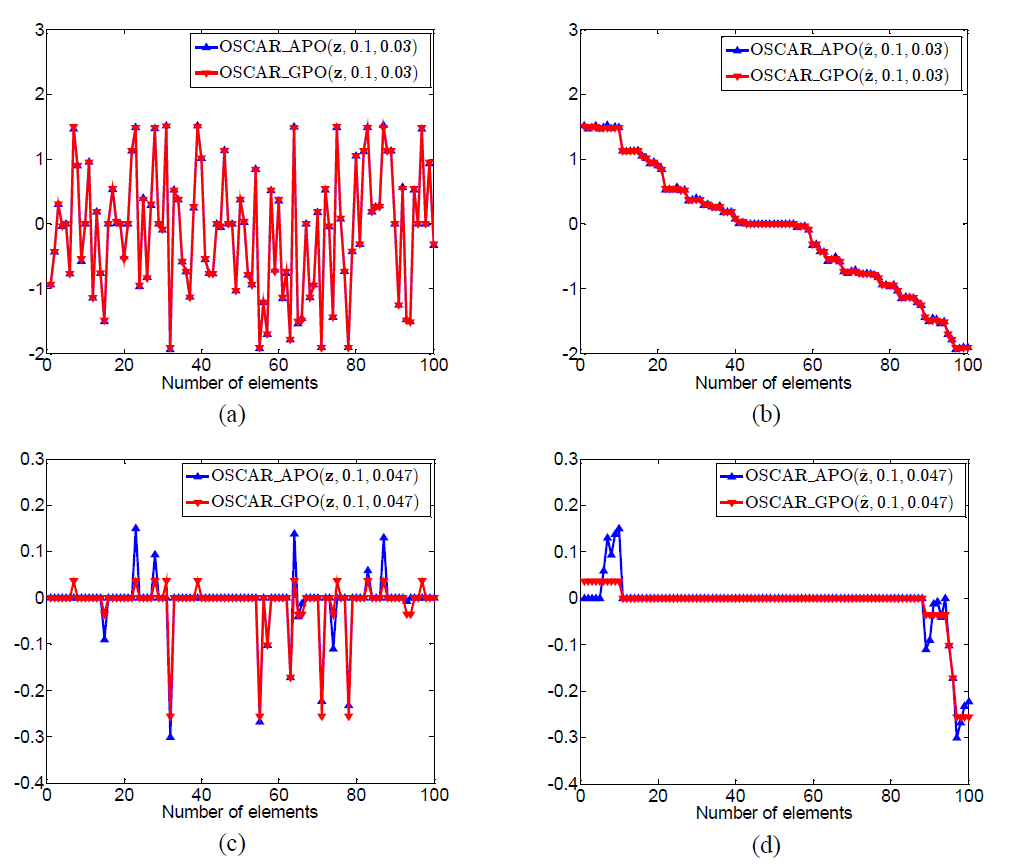}
	\caption{Comparisons on the results obtained by APO and GPO:
	(a) APO and GPO with ${\bf z}$, $\lambda_1=0.1$ and $\lambda_2=0.03$;
	(b) with $\hat{\bf z}$, $\lambda_1=0.1$ and $\lambda_2=0.03$;
	(c) with ${\bf z}$, $\lambda_1=0.1$ and $\lambda_2=0.047$;
	(d) with $\hat{\bf z}$, $\lambda_1=0.1$ and $\lambda_2=0.047$.}
	\label{fig:comparisonsAPOandGPO}
\end{figure}

To obtain further insight into the difference between APO and GPO,
we let $\hat{\bf z} = \hat{\bf P}({\bf z})\, {\bf z}$ be a
value-sorted version, \ie, such that $\hat{z}_i>\hat{z}_{i+1}$ for $i=1, \cdots, n-1$
 (note that $\hat{\bf z}$ is a value-sorted vector, while $\tilde{\bf z}$ above
is a magnitude-sorted vector).
 We also compare the results of
applying APO and GPO to ${\bf z}$ and $\hat{\bf z}$, for $\lambda_1 =  0.1$ and $\lambda_2 \in \{ 0.03,\, 0.047\}$;
the results are shown in Figure \ref{fig:comparisonsAPOandGPO}. We see that, for a smaller value of
$\lambda_2$, the result obtained by APO is more similar to those obtained by GPO,
and the fewer zeros are obtained by shrinkage. From Figure \ref{fig:comparisonsAPOandGPO} (b) and (d),
we conclude that the output of GPO on $\hat{\bf z}$ respects its decreasing value,
and exhibits the grouping effect via the averaging operation on the corresponding components,
which coincides with the analysis above.

\begin{figure}
	\centering
		\includegraphics[width=0.85\columnwidth]{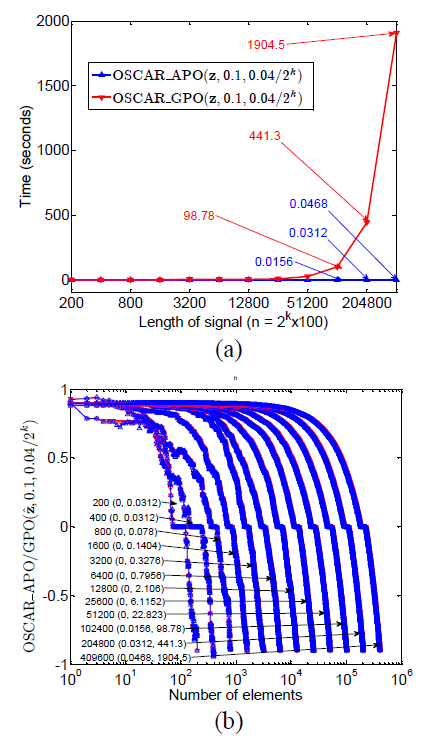}
	\caption{Speed comparisons of APO and GPO: (a) APO and GPO, with $\lambda_1=0.1$ and
    $\lambda_2=0.04/2^k, k\in {1, \cdots, 12}$; (b) APO and GPO operating on $\hat{\bf z}$
		(value-sorted vector), for	the same values of $\lambda_1$ and $\lambda_2$.
		The notation ``$\alpha \left(\beta,\gamma \right)$"
	represents ``\textbf{length of signal (Time of APO, Time of GPO)}". The
	horizontal axis corresponds to the signal length $n = 2^k \times 100, k\in {1, \cdots, 12}$
	 (different values of $\alpha$),
	while the vertical axis represents the APO and GPO of $\hat{\bf z}$ with
	different values of $\alpha$.}
	\label{fig:comparisonSpeed}
\end{figure}

We next compare APO and GPO in terms of CPU time. Signal ${\bf z}$ is randomly generated as
above, but now with length $n = 2^k \times 100$, for $k \in \left\{1, \cdots, 12\right\}$.
We set $\lambda_1=0.1$ and $\lambda_2=0.04/2^k $, for which $w_i \in \left[0.1, 4\right]$, for all
values of $k$. The results obtained by APO and GPO, with input signals ${\bf z}$ and $\hat{\bf z}$,
are shown in Figure \ref{fig:comparisonSpeed}. The results in this figure show that, as $n$ increases,
APO becomes clearly faster than GPO.  Figure \ref{fig:comparisonSpeed} (b) shows that as $n$ increases,
the results obtained by APO approach those of GPO.

\Section{Solving OSCAR problems using PSAs}\label{sec:OSCARandPSAS}

We consider the following general problem

\begin{equation}\label{generalproblem}
\min_{\bf x} h({\bf x}) :=  f({\bf x}) + g({\bf x})
\end{equation}
where $f,g: \mathbb{R}^{n}\rightarrow \mathbb{R} \cup \left\{-\infty, +\infty \right\}$
are convex functions (thus $h$ is a convex function), where $f$ is smooth with a Lipschitz
continuous gradient of constant $L$, while $g$ is possibly nonsmooth. OSCAR (see (\ref{OSCAR}))
is a special case of (\ref{generalproblem}) with $f({\bf x})=\frac{1}{2} \left\|{\bf y}-{\bf A}{\bf x}\right\|^2$,
and $g({\bf x})=r_{\mbox{\tiny OSCAR}}({\bf x})$. In this paper, we assume that (\ref{generalproblem}) is solvable, \ie,
the set of minimizers is not emply: $\arg\min_{{\bf x}} h({\bf x}) \neq \varnothing$.

To solve (\ref{generalproblem}), we investigate six state-of-the-art PSAs: FISTA \cite{beck2009fast},
TwIST \cite{bioucas2007new}, SpaRSA \cite{wright2009sparse}, ADMM \cite{boyd2011distributed},
SBM \cite{goldstein2009split} and PADMM \cite{chambolle2011first}. In each of these algorithms,
we apply APO and GPO. It is worth recalling that GPO can give the exact solution, while APO cannot,
so that the algorithms with GPO are exact ones, while those with APO are inexact ones.
Due to space limitation, we next only detail FISTA, SpaRSA, and PADMM applied to OSCAR; the
detailes of the other algorithms can be obtained from the corresponding publications. In our
experiments, we conclude that SpaRSA is the fastest algorithm and PADMM yields the best solutions.

\SubSection{FISTA}

FISTA is a fast version of the iterative shrinkage-thresholding (IST) algorithm, based
on Nesterov's acceleration scheme \cite{nesterov1983method}, \cite{nesterovintroductory}.
The FISTA algorithmic framework for OSCAR is as follows.

\vspace{0.3cm}
\begin{algorithm}{FISTA}{
\label{alg:fista}}
Set $k=0$, $t_1=1$, ${\bf u}_0$ and  compute $L$.\\
\qrepeat\\
     ${\bf v}_{k} = {\bf u}_k -{\bf A}^T \left({\bf A} {\bf u}_k  - {\bf y}\right)/L$\\
     ${\bf x}_{k}  = \mbox{ProxStep} \left({\bf v}_{k},\lambda_1/L,\lambda_2/L  \right)$\\
		 $t_{k+1}=\left(1 + \sqrt{1 + 4 t_k^2}\right)/2$ \\
		 $ {\bf u}_{k+1} = {\bf x}_{k} + \frac{t_k - 1}{t_{k+1}} \left({\bf x}_{k+1} - {\bf x}_{k}\right)$ \\
     $k \leftarrow k + 1$
\quntil some stopping criterion is satisfied.
\end{algorithm}
\vspace{0.3cm}

In the algorithm, the function $\mbox{ProxStep}$ is either the GPO or the APO defined above;
this notation will also be adopted in the algorithms described below. The FISTA with $\mbox{OSCAR\_APO}$ is
termed FISTA-APO, and with $\mbox{OSCAR\_GPO}$ is termed FISTA-GPO (which is equivalent to the algorithm
proposed in \cite{zhong2012efficient}).

\SubSection{SpaRSA}
SpaRSA \cite{wright2009sparse} is another fast variant of IST which gets its speed from using the
step-length selection method of Barzilai and Borwein \cite{barzilai1988two}.
Its application to OSCAR leads to the following algorithm~.

\vspace{0.3cm}
\begin{algorithm}{SpaRSA}{
\label{alg:sparsa}}
Set $k=1$, $\eta > 1$, $\alpha_{min}$, $0<\alpha_{\min}<\alpha_{\max}$,  and  ${\bf x}_0$.\\
     $\alpha_0 =\alpha_{min} $\\
     ${\bf v}_0 = {\bf x}_0 -{\bf A}^T \left({\bf A} {\bf x}_0 - {\bf y}\right)/\alpha_0$\\
     ${\bf x}_1  = \mbox{ProxStep} \left({\bf v}_0,\lambda_1/\alpha_0, \lambda_2/\alpha_0 \right)$\\
		 Set $k=1$ \\
\qrepeat\\
     ${\bf s}_k = {\bf x}_k - {\bf x}_{k-1}$ \\
		 $\hat{\alpha}_k = \frac{\left({\bf s}_k\right)^T{\bf A}^T{\bf A}{\bf s}_k}{\left({\bf s}_k\right)^T {\bf s}_k}$\\
		 $\alpha_k = \max\left\{\alpha_{\min}, \min\left\{\hat{\alpha}_k,\alpha_{\max}\right\}\right\}$\\
     \qrepeat\\
          ${\bf v}_k = {\bf x}_k -{\bf A}^T \left({\bf A} {\bf x}_k - {\bf y}\right)/\alpha_k$\\
          ${\bf x}_{k+1}  = \mbox{ProxStep}\left({\bf v}_k , \lambda_1/\alpha_k, \lambda_2/\alpha_k\right)$\\
		      $\alpha_k \leftarrow \eta \alpha_k$
		 \quntil ${\bf x}_{k+1}$ satisfies an acceptance criterion.\\
     $k \leftarrow k + 1$
\quntil some stopping criterion is satisfied.
\end{algorithm}
\vspace{0.3cm}

The typical acceptance criterion (in line 14) is to guarantee that the objective function
decreases; see \cite{wright2009sparse} for details. SpaRSA with $\mbox{OSCAR\_APO}$ is termed SpaRSA-APO,
and with $\mbox{OSCAR\_GPO}$ is termed SpaRSA-GPO.

\SubSection{PADMM}
PADMM \cite{chambolle2011first} is a preconditioned version of ADMM, also an efficient
 first-order primal-dual algorithm for convex optimization problems with known saddle-point structure.
 As for OSCAR using PADMM, the algorithm is shown as the following:

\vspace{0.3cm}
\begin{algorithm}{PADMM}{
\label{alg:padmm}}
Set $k=0$, choose $\mu$, ${\bf v}_{0}$ and ${\bf d}_{0}$.\\
\qrepeat\\
     ${\bf v}_{k+1} = \prox_{f^*/\mu} \left( {\bf v}_{k} + {\bf A} {\bf d}_{k}/ \mu \right)$\\
     ${\bf x}_{k+1}  = \mbox{ProxStep} \left({\bf x}_{k} - {\bf A}^T {\bf v}_{k+1}/\mu , \lambda_1/\mu , \lambda_2 / \mu\right)$\\
		 ${\bf d}_{k+1} = 2{\bf x}_{k+1} - {\bf x}_{k}$\\
     $k \leftarrow k + 1$
\quntil some stopping criterion is satisfied.
\end{algorithm}
\vspace{0.3cm}

In PADMM, $f^*$ is the conjugate of  $f$, and $\prox_{f^*/\mu}$ can be obtained from Moreau's
identity \eqref{moerauidentity}). PADMM with $\mbox{OSCAR\_APO}$ is termed PADMM-APO, and
with $\mbox{OSCAR\_GPO}$ is termed PADMM-GPO.

%

\SubSection{Convergence}
We now turn to the convergence of above algorithms with GPO and APO.
Since GPO is an exact proximity operator, the convergences of the algorithms with
GPO are guaranteed by their own convergence results. However, APO is an approximate one,
thus the convergence of the algorithms with APO is not mathematically clear,
and we leave it as an open problem here, in spite of that we have practically
found that APO behaves similarly as GPO when the regularization parameter
 $\lambda_2$ is set to a  small enough value.

\Section{Numerical Experiments} \label{sec:experiments}
We report results of experiments on group-sparse signal recovery aiming at showing the differences
among the six aforementioned PSAs, with GPO or APO..
All the experiments were performed using MATLAB on a 64-bit Windows 7 PC with an
Intel Core i7 3.07 GHz processor and 6.0 GB of RAM. In order to measure the performance
of different algorithms, we employ the following four metrics defined on an estimate
${\bf e}$ of an original vector ${\bf x}$:
\begin{itemize}
	\item Elapsed time (termed \textbf{Time});
	\item Number of iterations (termed \textbf{Iterations});
	\item Mean absolute error ($\textbf{MAE}=\left\|{\bf x} - {\bf e}\right\|_1/n$);
	\item Mean squared error ($\textbf{MSE}=\left\|{\bf x} - {\bf e}\right\|_2^2/n$).
\end{itemize}

The observed vector ${\bf y}$ is simulated by ${\bf y} = {\bf A}{\bf x} + {\bf n}$,
where ${\bf n} \thicksim \mathcal{N} \left( 0, \sigma^2I\right) $ and
with $\sigma = 0.4$, and the original signal ${\bf x}$ is a $1000$-dimensional vector
with the following structure:
\begin{multline}\label{orignalsignal}
{\bf x} = [
\underbrace{\alpha_1,\cdots,\alpha_{100}}_{100},  \underbrace{0,\cdots,0}_{200},
\underbrace{\beta_1,\cdots,\beta_{100}}_{100}, \\ \underbrace{0,\cdots,0}_{200},
\underbrace{\gamma_1,\cdots,\gamma_{100}}_{100},\underbrace{0,\cdots,0}_{300}]^T
\end{multline}
where $\alpha_k = 7 + \epsilon_k$, $\beta_k = 9 + \epsilon_k$,
 $\gamma_k = -8 + \epsilon_k$ and $\epsilon_k \thicksim \mathcal{N} \left( 0, 1 \right) $,
the disturbance $\epsilon_k$ in each group can somehow reflect the real applications,
and ${\bf x}$ possesses both positive and negative groups. The  sensing matrix
${\bf A}$ is a $500 \times 1000$ matrix whose components are sampled from the standard normal distribution.

\begin{figure}
	\centering
		\includegraphics[width=0.9\columnwidth]{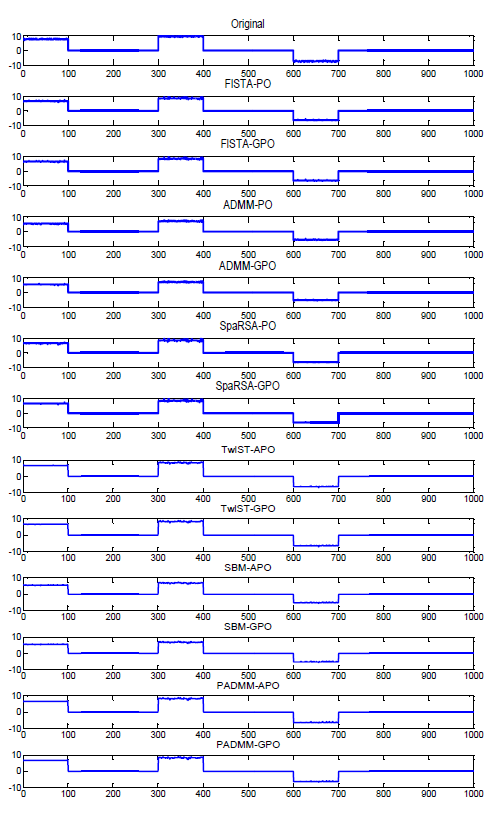}
	\caption{Recovered group-sparse signals.}
	\label{fig:RecoveredSignal}
\end{figure}

\begin{figure}
	\centering
		\includegraphics[width=0.9\columnwidth]{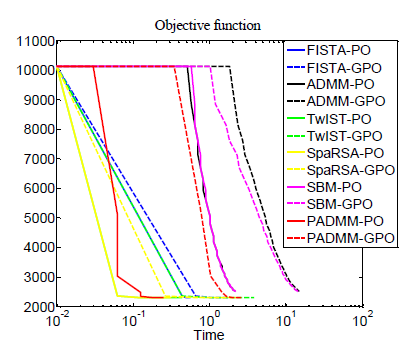}
	\caption{Objective function evolution over time.}
	\label{fig:objective}
\end{figure}

\begin{figure}
	\centering
		\includegraphics[width=0.8\columnwidth]{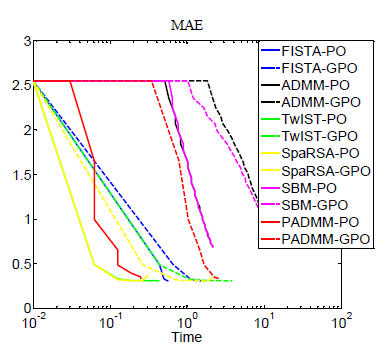}
	\caption{MAE evolution over time.}
	\label{fig:MAE}
\end{figure}

\begin{figure}
	\centering
		\includegraphics[width=0.8\columnwidth]{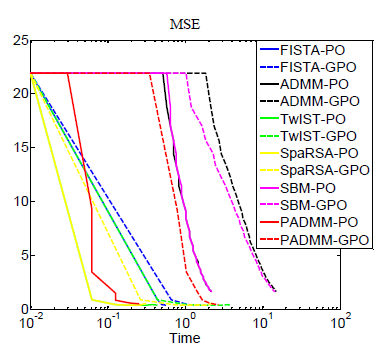}
	\caption{MSE evolution over time.}
	\label{fig:MSE}
\end{figure}

\SubSection{Recovery of Group-Sparse Signals}
We ran the aforementioned twelve algorithms: FISTA-GPO, FISTA-APO, TwIST-GPO, TwIST-APO,
SpaRSA-GPO, SpaRSA- APO, SBM-GPO, SBM-APO, ADMM-GPO, ADMM-APO, PADMM-GPO and PADMM-APO.
The stopping condition is $\left\|{\bf x}_{k+1}-
 {\bf x}_k\right\|/\left\|{\bf x}_{k+1}\right\|\leq tol$, where $tol = 0.01$ and ${\bf x}_k$
represents the estimate at the $k$-th iteration. In the OSCAR model, we
set $\lambda_1=0.1$ and $\lambda_2=0.001$. The original and recovered signals are shown in
Figure \ref{fig:RecoveredSignal}, and the results of Time, Iterations, MAE, and MSE are shown in
Table \ref{tab:resultsofsparserecovery}.
Figure \ref{fig:objective}, \ref{fig:MAE} and \ref{fig:MSE} show the evolution of
the objective function, MAE, and MSE over time, respectively.
\begin{table}[t]
\centering \caption{RESULTS OF GROUP-SPARSE SIGNAL RECOVERY} \label{tab:resultsofsparserecovery}
\begin{tabular}{|l|l|l| l|l|}
\hline\rule[-0.1cm]{0cm}{0.4cm}
 &{\footnotesize Time (seconds)} & {\footnotesize Iterations}  & {\footnotesize MAE} &  {\footnotesize MSE}\\
\hline
\footnotesize FISTA-APO & \footnotesize 0.562 &  \footnotesize 4   & \footnotesize 0.337 & \footnotesize 0.423  \\
\footnotesize FISTA-GPO   & \footnotesize 1.92 & \footnotesize 4   & \footnotesize 0.337  & \footnotesize 0.423 \\
\footnotesize TwIST-APO & \footnotesize 0.515  & \footnotesize 7    & \footnotesize 0.342  & \footnotesize 0.423  \\
\footnotesize TwIST-GPO   & \footnotesize 4.45  & \footnotesize 7    & \footnotesize 0.342  & \footnotesize 0.423  \\
\footnotesize SpaRSA-APO & \footnotesize 0.343  & \footnotesize 6  & \footnotesize 0.346  & \footnotesize 0.445 \\
\footnotesize SpaRSA-GPO   & \footnotesize 2.08  & \footnotesize 6   & \footnotesize 0.375  & \footnotesize 0.512  \\
\footnotesize ADMM-APO & \footnotesize 2.23  & \footnotesize 37    & \footnotesize 0.702  & \footnotesize 1.49   \\
 \footnotesize ADMM-GPO & \footnotesize 16.2 & \footnotesize 37    & \footnotesize 0.702  & \footnotesize 1.49 \\
\footnotesize SBM-APO & \footnotesize 2.26 & \footnotesize 37    & \footnotesize 0.702  & \footnotesize 1.49 \\
\footnotesize SBM-GPO & \footnotesize 14.9 & \footnotesize 37   & \footnotesize 0.702  & \footnotesize 1.49  \\
\footnotesize PADMM-APO & \footnotesize 0.352  & \footnotesize 9  & \footnotesize 0.313  & \footnotesize 0.411  \\
\footnotesize PADMM-GPO & \footnotesize 2.98  & \footnotesize 9     & \footnotesize 0.313  & \footnotesize 0.411  \\
\hline
\end{tabular}
\end{table}

We can conclude from Figures \ref{fig:objective}, \ref{fig:MAE}, and \ref{fig:MSE}, and
Table \ref{tab:resultsofsparserecovery}, that PSAs with APO are much faster than that with
GPO. Among the PSAs with GPO and APO, SpaRSA-APO is the fastest one in time,
while FISTA-APO uses fewer iterations. The results obtained by PADMM are the most
accurate ones.

\SubSection{Impact of Signal Length}

\begin{figure}
	\centering
		\includegraphics[width=0.80\columnwidth]{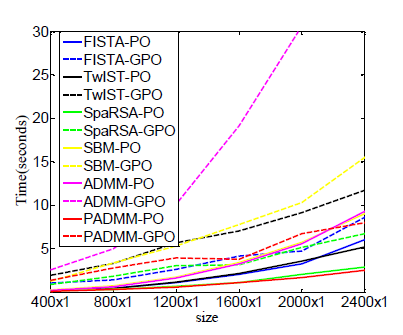}
	\caption{Consumed time of different algorithms over signal length.}
	\label{fig:SensitiveSize}
\end{figure}

Finally, we study the influence of the signal length on the performance of the algorithms. For
signal length $n$, we use a matrix A of size $\frac{n}{2} \times n$, and keep other setup
of above subsection unchanged.  The results are shown in Figure
\ref{fig:SensitiveSize}, where the horizontal axis represents the signal size.
From Figure \ref{fig:SensitiveSize}, we can conclude that, along with increasing signal size,
the speed of the PSAs with APO is much faster than that with GPO such that
the former are more suitable for solving large-scale problems.

\Section{Conclusions}
\label{sec:conclusions}

We have proposed approaches to efficiently solve problems involving the OSCAR regularizer, which outperforms other group-sparsity-inducing ones.
The exact and approximate versions of the proximity operator of the OSCAR regularizer were considered, and their differences were analyzed both mathematically and numerically. Naturally, the approximate is faster than the exact one, but for certain range of the parameters of the regularizer, the results are very similar. These two proximity operators provide a very useful building bock for
the applications of proximal spitting algorithms. We have considered six state-of-the-art algorithms: FISTA, TwIST, SpaRSA, SBM, ADMM and PADMM. 
Experiments on group-sparse signal recovery have shown that these algorithms, working with the approximate proximity operator, are able to obtain accurate
results very fast. However, their mathematical convergence proof is left as an 
open problem, whereas the algorithms operating with the exact proximity operator inherit the corresponding convergence results.

\section{Acknowledgements}

We thank James T. Kwok for kindly providing us the C++ code of their paper \cite{zhong2012efficient}.

\bibliographystyle{IEEEtranS}
\bibliography{bibfile}

\end{document}